\pdfoutput=1

\documentclass[11pt]{article}

\usepackage{acl}

\usepackage{times}
\usepackage{latexsym}

\usepackage[T1]{fontenc}

\usepackage[utf8]{inputenc}

\usepackage{microtype}

\usepackage{graphics}
\usepackage{graphicx}
\usepackage{amsmath}
\usepackage{booktabs}
\usepackage{csquotes}
\usepackage[disable]{todonotes}

\title{Social-Group-Agnostic Word Embedding Debiasing \\via the Stereotype Content Model}

\author{Ali Omrani \and Brendan Kennedy \and Mohammad Atari \and
Morteza Dehghani \\
University of Southern California\\
\texttt {\{aomrani,btkenned,atari,mdehghan\}@usc.edu}}

\begin{document}
\maketitle
\begin{abstract}
Existing word embedding debiasing methods require \textit{social-group-specific} word pairs (e.g., ``man''--``woman'') for each social attribute (e.g., gender), which cannot be used to mitigate bias for other social groups, 
making these methods impractical or costly to incorporate understudied social groups in debiasing. 
We propose that the Stereotype Content Model (SCM), 
a theoretical framework developed in social psychology for understanding the content of stereotypes, which structures stereotype content along two psychological dimensions --- ``warmth'' and ``competence'' --- can help debiasing efforts to become \textit{social-group-agnostic} by capturing the underlying connection between bias and stereotypes. Using only pairs of terms for warmth (e.g., ``genuine''--``fake'') and competence (e.g., ``smart''--``stupid''), we perform debiasing with established methods and find that, across gender, race, and age, SCM-based debiasing performs comparably to group-specific debiasing. 
\end{abstract}
\section{Introduction}

The societal impacts of Natural Language Processing (NLP) have stimulated research on measuring and mitigating the unintended social-group biases encoded in language models \cite{hovy2016social}. However, the majority of this important line of work is atheoretical in nature and ``fails to engage critically with what constitutes `bias' in the first place'' \cite{blodgett2020language}. Although there is a multitude of approaches to word embedding debiasing \citep{bolukbasi2016man,zhao2018learning,dev2019attenuating}, most of these approaches rely on group-specific bias subspaces. Recently, \citet{gonen2019lipstick} found that, for example, gender bias in word embeddings is more systematic than simply debiasing along the ``gender'' subspace. 
Combining well-established models of bias and stereotyping from social psychology with word embedding debiasing efforts, we follow \citet{blodgett2020language} in proposing a theory-driven debiasing approach that does not rely on a particular social group such as gender or race.
We show that, by relying on a theoretical understanding of social stereotypes to define a group-agnostic bias subspace, word embeddings can be adequately debiased across multiple social attributes.


Group-specific debiasing, which debiases along subspaces defined by social groups (e.g., gender or race), is not only atheoretical but also unscalable. For example, previous works' excessive focus on gender bias in word embeddings has driven the development of resources for  gender debiasing (e.g., equality sets for gender), but biases associated with other social groups remain understudied. This is due to the fact that resources developed for one social group (e.g., gender) do not translate easily to other groups. 
Furthermore, group-specific debiasing is limited in terms of generalizability: as shown by  \citet{agarwal2019word}, stereotype content in word embeddings is deep-rooted, and thus is not easily removed using explicit sets of group-specific words. In contrast, a social-group-agnostic approach would not have such restrictions.

Social-group bias mitigation, as a societal problem, can benefit from social psychological theories to understand the underlying structure of language-embedded biases rather than attending to ad hoc surface patterns.
The Stereotype Content Model (SCM) \cite{fiske2002model} is a theoretical framework developed in social psychology to understand the content and function of stereotypes in interpersonal and intergroup interactions. The SCM proposes that human stereotypes are captured by two primary dimensions of \textit{warmth} (e.g., trustworthiness, friendliness) and \textit{competence} (e.g., capability, assertiveness). From a socio-functional, pragmatic perspective, people's perception of others' intent (i.e., warmth) and capability to act upon their intentions (i.e., competence) affect their subsequent emotion and behavior \cite{cuddy2009stereotype}. Depending on historical processes, various social groups may be located in different stereotypic quadrants (high vs. low on warmth and competence) based on this two-dimensional model. 

Here, we propose that SCM-based debiasing can provide a theory-driven and scalable solution for mitigating social-group biases in word embeddings.  
In our experiments, we find that by debiasing with respect to the subspace defined by warmth and competence, our SCM-based approach performs comparably with group-specific debiasing. 
Our approach fares well both in terms of bias reduction and the preservation of embedding utility (i.e., the preservation of semantic and syntactic information) \citep{bolukbasi2016man}, while having the advantage of being social-group-agnostic.

\section{Background}
\subsection{Post hoc Word Embedding Debiasing}

Our work builds on post hoc debiasing, removing biases by modifying pre-trained word embeddings. 
Most work we review focuses on gender-related debiasing \cite[e.g.,][]{bolukbasi2016man, zhao2018learning, dev2019attenuating}; importantly, we focus our work on other social categories as well, bringing attention to these understudied groups.

Originally, \citet{bolukbasi2016man} proposed Hard Debiasing (HD) for gender bias. HD removes the gender component from inherently non-gendered words and enforces an equidistance property for inherently gendered word pairs (equality sets).
Two follow-ups to this work include \citet{manzini2019black}, which formulated a multiclass version of HD for attributes such as race, and 
\citet{dev2019attenuating}, which introduced Partial Projection, a method that does not require equality sets and is more effective than HD in reducing bias. Extending these approaches to other social attributes is not trivial because a set of definitional word pairs has to be curated for each social group, which is a dynamic and context-dependent task because these pairs are dependent on historical moment.  

\citet{gonen2019lipstick} provided evidence that gender bias in word embeddings is deeper than previously thought, and methods based on projecting words onto a ``gender dimension'' only hide bias superficially. They showed that after debiasing, most words maintain their relative position in the debiased subspace. 
In this work, we address the shortcomings highlighted by \citeauthor{gonen2019lipstick} and \citeauthor{agarwal2019word} with a theory-driven bias subspace, rather than algorithmic improvement.
  
\subsection{Bias and the Stereotype Content Model}
The bias found in language models is rooted in human biases \cite{caliskan2022social}; thus, to alleviate such biases, we should ground our debiasing approaches in social psychological theories of stereotyping \cite{blodgett2020language}. The Stereotype Content Model (SCM) \cite{fiske2002model, cuddy2009stereotype} is a social psychological theory positing that stereotyping of different social groups can be captured along two orthogonal dimensions, 
\enquote{warmth} and \enquote{competence.} The warmth dimension of stereotypes has to do with people's intentions in interpersonal interactions, while the competence dimension has to do with assessing others' ability to act on those intentions. 
While there are a number of other social psychological theories capturing outgroup biases \citep[e.g.,][]{zou2017two,koch2016abc}, SCM has been shown to predict emotional and behavioral reactions to societal outgroups.

\subsection{The SCM and Language}
SCM is a well-established theoretical frameworks of stereotyping, and has begun to be applied in NLP. 
Recently \citet{nicolas2021comprehensive} developed dictionaries to measure warmth and competence in textual data. Each dictionary was initialized with a set of seed words from the literature which was further expanded using WordNet \cite{miller1995wordnet} to increase the coverage of stereotypes collected from a sample of Americans. \citet{fraser2021understanding} showed that, in word embeddings, SCM dictionaries capture the group stereotypes documented in social psychological research. 
Recently, \citet{mostafazadeh2021hate} applied SCM dictionaries to quantify social group stereotypes embedded in language, demonstrating that patterns of prediction biases can be explained using social groups' warmth and competence embedded in language.


\section{Methods \& Evaluation}
There are two components to each post hoc debiasing approach: the \textbf{Bias Subspace}, which determines the subspace over which the algorithms operate, and the \textbf{Algorithm}, which is how the word embeddings are modified with respect to the bias subspace. In this section, we review the concept of bias subspaces, established algorithms for debiasing, and how bias is quantified in word embeddings. Finally, we introduce our social-group-agnostic framework; SCM-based debiasing.  

\subsection{Identifying a Bias Subspace}
\label{sec:bias_subspace}
Post hoc word embedding bias mitigation algorithms operate over a subspace of bias in the embedding space. 
Given a set $D = \{(d^+_1,d^-_1),... ,(d^+_n,d^-_n)\}$ of word pairs that define the bias concept (e.g. ``father''--``mother'' for gender) 
the bias subspace $v_B$ is the first $k$ principal components of matrix $C$, constructed from stacking the difference in embeddings of $d^+_i$ and $d^-_i$. 

\subsection{Debiasing Algorithms}
\label{sec:debias_algos}
Method definitions below use the following notation: $W$ denotes vocabulary, $\vec{w}$ and $\vec{w'}$ denote the embedding of word $w$ before and after debiasing.

\smallskip\noindent\textbf{Hard Debiasing (HD)~~}
An established approach for mitigating bias in word embeddings is Hard Debiasing \citep[HD;][]{bolukbasi2016man}. For gender, HD removes the gender subspace from words that are not inherently gendered by projecting them orthogonal to gender subspace. For word pairs that are inherently gendered, HD equalizes them, modifying the embeddings such that they are equidistant from the inherently non-gendered words. 





\smallskip\noindent\textbf{Subtraction (Sub) ~~}
Subtraction (Sub) was introduced as a baseline by \citet{dev2019attenuating} wherein the bias subpspace $v_B$ is subtracted from all word vectors. Formally, for all $w \in W$, $\vec{w'} := \vec{w} - \vec{v_B}$.

\smallskip\noindent\textbf{Linear Projection (LP)~~}
To mitigate the bias with respect to bias dimension $v_B$, Linear Projection (LP) projects every word $w \in W$ to be  orthogonal to $v_B$. Formally, $\vec{w'} := \vec{w} - <\vec{w},\vec{v_B}> \vec{v_B}$.

\smallskip\noindent\textbf{Partial Projection (PP)~~}
To improve on LP, Partial Projection (PP) was developed to allow the extent of projection to vary based on the component of the given word vector which is orthogonal to the bias subspace. Intuitively, 
only words with unintended bias (e.g., ``nurse'' or ``doctor''), and not words which are definitional to the bias concept (e.g., ``man'' or ``woman'') will have a large orthogonal component to the bias subspace $v_B$. 
For all words $w \in W$, 
\begin{align*}
    w' &= \mu + r(w) + \beta \cdot f(\|r(w)\|)\cdot v_B\\
    \beta &= \langle w, v_B\rangle - \langle\mu, v_B\rangle
\end{align*}
where $\mu$ is the mean embedding of words used to define $v_B$, $r(w) = w - \langle w,v_B\rangle v_B $ is the bias-orthogonal component, and $f$ is a smoothing function, for example $f(\eta) = \frac{\sigma ^ 2}{(\eta + 1)^2}$, which helps to remove unintended bias and keep definitional bias \citep[see][]{dev2019attenuating}. 





\subsection{Measures of Bias in Word Embeddings}
\label{sec:bias_measures}

\smallskip\noindent\textbf{Embedding Coherence Test~~}
Given a list of word pairs $A =  \{(a^+_{1},a^-_{1}), ..., (a^+_{k},a^-_{k})\}$, indicating two ``poles'' of a social attribute, and a set of professions $P = \{p_1, ..., p_m\}$, the Embedding Coherence Test \cite[ECT;][]{dev2019attenuating} 
is the Spearman rank correlation between the rank order of cosine similarities of professions with each pole's average embedding.
Ideally, if bias is completely removed, poles should achieve identical ordering of associations with professions (ECT $= 1$).


\smallskip\noindent\textbf{Embedding Quality Test~~} The EQT \citep{dev2019attenuating} quantifies the improvement in unbiased analogy generation after debiasing. Similar to ECT, EQT requires a set of word pairs $A$ and a set of professions $P$. For each word pair $(a^+_{i},a^-_{i})$ the analogy $a^+_{i}:a^-_{i}::p_j$ is completed, if the answer is $p_j$ or plurals or synonyms of $p_j$ (via NLTK; \citealt{bird2009natural}), it is counted as unbiased. EQT is the ratio of unbiased analogies to all analogies. An ideal unbiased model would achieve EQT$=1$ while lower values indicate a more biased model. 

\subsection{SCM-Based Debiasing}
To identify a \textit{group-agnostic} bias subspace, we use the warmth and competence dictionaries from \citep{nicolas2021comprehensive}.
To construct the poles of the dimensions, ``high'' and ``low'' word pairs (e.g., ``able''--``unable'' for competence and ``sociable''--``unsociable'' for warmth) were selected by down-sampling to 15 word pairs, per dimension.
We use word pairs for each SCM dimension to identify an SCM subspace (see Section~\ref{sec:bias_subspace}), and subsequently apply the methods from Sec.~\ref{sec:debias_algos}.

\section{Experiments}

\begin{table*}[!ht]
    \centering
    \begin{tabular}{|l|c|c|cc|cc|cc|c|}
    \hline
         & \multicolumn{1}{|c}{Vanilla} & HD$_\text{same}$ & Sub$_\text{same}$ & Sub$_\text{SCM}$ & LP$_\text{same}$ & LP$_\text{SCM}$ &  PP$_\text{same}$ & PP$_\text{SCM}$ & PP$_\text{G+R+A}$\\\hline
        ECT$_\text{gender}$  & 0.83 & 0.92 & \textbf{0.83} &  \textbf{0.83} & 0.82 &  \textbf{0.83} & \textbf{0.99} & 0.97 & 0.99  \\
        ECT$_\text{race}$    & 0.69 & -    & \textbf{0.51} &  \textbf{0.52}   & 0.70  &  \textbf{0.74}    & \textbf{0.99} & 0.96 & 0.99\\
        ECT$_\text{age}$     & 0.30 & -    & 0.23 &   \textbf{0.34}  & \textbf{0.60}  &  0.34   & \textbf{0.96} & 0.95 & 0.99  \\
        \hline
        EQT$_\text{gender}$ & 0.075 & 0.056 & \textbf{0.071} &  \textbf{0.072}   & \textbf{0.081} & 0.073  & \textbf{0.063} & 0.049 & 0.059  \\
        EQT$_\text{race}$ & 0.042 & -       & 0.032 &  \textbf{0.036}  & \textbf{0.051} & 0.044    & \textbf{0.061} & 0.056 & 0.073\\
        EQT$_\text{age}$ & 0.052 & -        & \textbf{0.043} &  0.041   & \textbf{0.062} & 0.051   & \textbf{0.063} & 0.047 & 0.057\\
        \hline
    \end{tabular}
    \caption{ECT and EQT for gender, race, and age. 
    Debiasing was repeated 30 times for each method, and bold values indicate higher scores (per method) with non-overlapping 95\% confidence intervals. HD was limited to gender because of other dimensions' lack of equality sets.}
    \label{tab:ect-eqt}
    \vspace{-3mm}
\end{table*}

We test whether SCM-based debiasing can substitute for group-specific debiasing simultaneously for gender, race, and age.
This is broken down into two related research questions. First, does SCM-based debiasing remove a comparable amount of bias relative to group-specific debiasing? And second, does SCM-based debiasing have more or less of a negative effect on embedding utility \citep{bolukbasi2016man}? 
We compare SCM-based debiasing to group-specific debiasing using previous debiasing methods, specifically HD, Sub, LP, and PP (Section~\ref{sec:debias_algos}), and evaluate bias as measured by ECT and EQT following \citet{dev2019attenuating}. 
In addition, we evaluate the performance of each set of debiased embeddings on established word embedding benchmarks \cite{jastrzebski2017evaluate}. 

\subsection{Bias Reduction}
\label{eval_debiasing}

We investigate whether SCM-based debiasing can simultaneously debias word embeddings with respect to gender, race, and age.
For a given bias dimension, we established baselines by applying HD, Sub, LP, and PP using the respective word pair list (e.g., for gender bias we used gender word pairs), denoted with the subscript ``same.'' To place an upper bound on removed bias, we perform PP using gender, race, and age word lists (PP$_\text{G+R+A}$). For race and age we used the lists from \citet{caliskan2017semantics}, while gender lists were taken from \citet{bolukbasi2016man}. 
All methods were repeatedly applied using 30 different word pair samples, and we report each measure's average and compare values using 95\% confidence intervals. 
Implementation details are provided in the Appendix.

Table~\ref{tab:ect-eqt} shows the results of our experiments. 
Overall, SCM-based debiasing performs comparably to social-group-specific debiasing across methods. Specifically for ECT, SCM-based debiasing was either better than, or not statistically different from, LP$_\text{same}$ and Sub$_\text{same}$, while SCM-based debiasing was only slightly out-performed by PP$_\text{same}$ ($0.01$--$0.03$). 
In other words, these results demonstrate that warmth and competence dimensions can simultaneously capture gender, race, and age bias in word embeddings. 
For the EQT, results are somewhat similar to those of ECT; however, we caution against interpreting small differences in EQT due to its definition of biased analogies relying on NLTK to compile comprehensive sets of synonyms and plural forms of words \citep{dev2019attenuating}. 

\subsection{Word Embedding Utility} \label{sec:utility}




\begin{table}[!htb]
    \centering
    \resizebox{\linewidth}{!}{%
    \begin{tabular}{lcccc}
         & \multicolumn{2}{c}{Analogy} & \multicolumn{2}{c}{Similarity} \\\cline{2-5}
         & \multicolumn{1}{c}{Google}  & MSR & WS353 & RG-65 \\\hline
        Vanilla                         & 0.39 & 0.45 & 0.50 & 0.50 \\\hline
        PP$_\text{Gender}$ & 0.31 & 0.36 & 0.49 & 0.37 \\
        PP$_\text{Gender+Race}$         & 0.27 & 0.33 & 0.43 & 0.30 \\
        PP$_\text{G+R+A}$     & 0.25  & 0.30 & 0.40 & 0.27\\
        PP$_\text{SCM}$                 & 0.29 & 0.34 & 0.42 & 0.33 \\\hline
        
    \end{tabular}
    }
    \caption{Embedding quality for debiased models.}
    \label{tab:embedding_quality}
    \vspace{-5mm}
\end{table}

Table~\ref{tab:ect-eqt} shows that PP$_\text{G+R+A}$ outperformed all other methods. However, one trade-off is the reduction in word embedding utility. Table~\ref{tab:embedding_quality} shows that PP$_\text{SCM}$ (PP applied to Warmth and Competence) preserves more embedding utility than PP$_\text{G+R+A}$, using established benchmarks for analogy and similarity \cite{jastrzebski2017evaluate}. Due to the information removed in the debiasing process, as the number of social attributes increases, the quality of embeddings for group-specific debiasing deteriorates; however, this is not the case for PP$_\text{SCM}$.
\section{Conclusion}

We demonstrated that social-group biases in word embeddings can be adequately mitigated in a social-group-agnostic way by operating along the SCM dimensions of warmth and competence \citep{fiske2002model}, introducing a new theory-driven approach for mitigating such biases. 
Our work shows that the SCM subspace can be used to mitigate bias for additional social groups without lowering word embedding quality.
Future directions of our work include: (1) using other social psychological frameworks for quantifying language-embedded social stereotypes \citep[e.g.,][]{koch2016abc}; (2) extending these findings from English to other languages \cite{kuvcera2022beyond}; and (3) the extension of our findings to other social group biases (e.g., nationality) \cite{herold2022applying} and intersectional biases (e.g., intersection of race and gender).
\section{Limitations}
    

Some limitations of the present work are worth noting. First, we note the contextual limitations of our current analysis using the proposed theory-driven framework. Specifically, the word embeddings used in this work are trained on contemporary English language and our social context overly contains explicit stereotypes encoded in English word embedding model. Stereotypes for the same group can be quite different, however, depending on the language and culture. 
While out of scope of the present work, cross-societal differences in human stereotyping have been shown to be explainable using the SCM framework \citep{cuddy2009stereotype}. Thus, it is possible that our SCM-based framework generalizes to social group biases beyond those in English. Future research is encouraged to replicate our study in non-English languages. 

Another limitation of our work is that we only test our proposed framework on word embeddings. We acknowledge that to maximize the impact of bias mitigation efforts, these methods need to be extended to the state-of-the-art language models. Future studies are encouraged to address this limitation. Be that as it may, we emphasize that our proposed approach is the first \textit{theory-driven} and generalizable approach in mitigating such social biases based on social psychological theories of stereotyping and bias. 

Furthermore, we would like to point out that there exists a catalogue of bias measurements for word embeddings in the field. Some of these measures have been shown to fail robustness checks. Although our current work uses some of the most recently developed ECT and EQT, we believe that few, if any, of these measurements are completely sound. Indeed, developing a new bias measurement scale is not within the scope of this work.



\bibliographystyle{acl_natbib}
\bibliography{references}

\clearpage

\end{document}